\documentclass[sigconf]{acmart}

\usepackage[utf8]{inputenc} % allow utf-8 input
\usepackage[T1]{fontenc}    % use 8-bit T1 fonts
\usepackage{hyperref}       % hyperlinks
\usepackage{url}            % simple URL typesetting
\usepackage{booktabs}       % professional-quality tables
\usepackage{amsfonts}       % blackboard math symbols
\usepackage{nicefrac}       % compact symbols for 1/2, etc.
\usepackage{microtype}      % microtypography

\usepackage{amsmath}
\usepackage{subcaption}
\usepackage{graphicx}
\usepackage{tikz}
\def\checkmark{\tikz\fill[scale=0.35](0,.35) -- (.25,0) -- (1,.7) -- (.25,.15) -- cycle;}
\usepackage{etoolbox}
\usepackage[ruled]{algorithm2e}
\usepackage{multirow}
\usepackage{hhline}
\usepackage{wrapfig}
\usepackage{comment}

\usepackage{xcolor}

\usepackage{rotating}

\usepackage{xspace}
\usepackage{units}

\definecolor{myred}{rgb}{0.8,0,0}
\definecolor{mygreen}{rgb}{0,0.6,0}
\definecolor{myblue}{rgb}{0,0,0.7}

\newcommand{\mujoco}{{\sc MuJoCo}\xspace}
\newcommand{\brax}{{\sc Brax}\xspace}
\newcommand{\jax}{{\sc Jax}\xspace}

\newcommand{\ant}{{\sc ant-v2}\xspace} 
\newcommand{\antmaze}{{\sc ant-maze}\xspace} 
\newcommand{\ptmaze}{{\sc point-maze}\xspace}
 
\newcommand{\anttrap}{{\sc ant-trap}\xspace}
\newcommand{\humanoidtrap}{{\sc humanoid-trap}\xspace}

\newcommand{\qdpg}{{\sc qd-pg}\xspace}

\newcommand{\tddd}{{\sc td3}\xspace} 
\newcommand{\sac}{{\sc sac}\xspace} 
\newcommand{\cemrl}{{\sc cem-rl}\xspace} 
\newcommand{\cem}{{\sc cem}\xspace} 
 
\newcommand{\pssstddd}{{\sc p3s-td3}\xspace}
\newcommand{\agac}{{\sc agac}\xspace}
\newcommand{\diayn}{{\sc diayn}\xspace}

\newcommand{\me}{{\sc map-elites}\xspace}

\newcommand{\pgame}{{\sc pga-me}\xspace}

\newcommand{\mees}{{\sc me-es}\xspace}

\newcommand{\rnd}{{\sc rnd}\xspace}

\newcommand{\nsres}{{\sc nsr-es}\xspace} 
\newcommand{\nsraes}{{\sc nsra-es}\xspace}

\AtBeginDocument{%
  \providecommand\BibTeX{{%
    \normalfont B\kern-0.5em{\scshape i\kern-0.25em b}\kern-0.8em\TeX}}}

\copyrightyear{2022}
\acmYear{2022}
\setcopyright{acmlicensed}\acmConference[GECCO '22 Companion]{Genetic and Evolutionary Computation Conference Companion}{July 9--13, 2022}{Boston, MA, USA}
\acmBooktitle{Genetic and Evolutionary Computation Conference Companion (GECCO '22 Companion), July 9--13, 2022, Boston, MA, USA}
\acmPrice{15.00}
\acmDOI{10.1145/3520304.3534038}
\acmISBN{978-1-4503-9268-6/22/07}

\begin{document}

\title{Assessing Quality-Diversity Neuro-Evolution Algorithms Performance in Hard Exploration Problems}


\author{Felix Chalumeau}
\affiliation{%
  \institution{InstaDeep}
  \city{Paris}
  \country{France}
}
\email{f.chalumeau@instadeep.com}

\author{Thomas Pierrot}
\affiliation{%
  \institution{InstaDeep}
  \city{Paris}
  \country{France}
}
\email{t.pierrot@instadeep.com}

\author{Valentin Mac\'e}
\affiliation{%
  \institution{InstaDeep}
  \city{Paris}
  \country{France}
}
\email{v.mace@instadeep.com}

\author{Arthur Flajolet}
\affiliation{%
  \institution{InstaDeep}
  \city{Paris}
  \country{France}
}
\email{a.flajolet@instadeep.com}

\author{Karim Beguir}
\affiliation{%
 \institution{InstaDeep}
 \city{London}
 \country{United Kingdom}
}
\email{kb@instadeep.com}

\author{Antoine Cully}
\affiliation{%
  \institution{Imperial College London}
  \city{London}
  \country{United Kingdom}
}
\email{a.cully@imperial.ac.uk}

\author{Nicolas Perrin-Gilbert}
\affiliation{%
  \institution{CNRS, Sorbonne Universit\'e}
  \city{Paris}
  \country{France}
}
\email{perrin@isir.upmc.fr}

\renewcommand{\shortauthors}{Chalumeau et al.}

\begin{abstract}
    A fascinating aspect of nature lies in its ability to produce a collection of organisms that are all high-performing in their niche. Quality-Diversity (QD) methods are evolutionary algorithms inspired by this observation, that obtained great results in many applications, from wing design to robot adaptation. Recently, several works demonstrated that these methods could be applied to perform neuro-evolution to solve control problems in large search spaces. In such problems, diversity can be a target in itself. Diversity can also be a way to enhance exploration in tasks exhibiting deceptive reward signals. While the first aspect has been studied in depth in the QD community, the latter remains scarcer in the literature. Exploration is at the heart of several domains trying to solve control problems such as Reinforcement Learning and QD methods are promising candidates to overcome the challenges associated. Therefore, we believe that standardized benchmarks exhibiting control problems in high dimension with exploration difficulties are of interest to the QD community. In this paper, we highlight three candidate benchmarks and explain why they appear relevant for systematic evaluation of QD algorithms. We also provide open-source implementations \footnote{\url{https://github.com/adaptive-intelligent-robotics/QDax}} in Jax allowing practitioners to run fast and numerous experiments on few compute resources.
\end{abstract}


%\ccsdesc[500]{Computer systems organization~Embedded systems}
%\ccsdesc{Computer systems %organization~Robotics}


\begin{teaserfigure}
        \includegraphics[width=\textwidth]{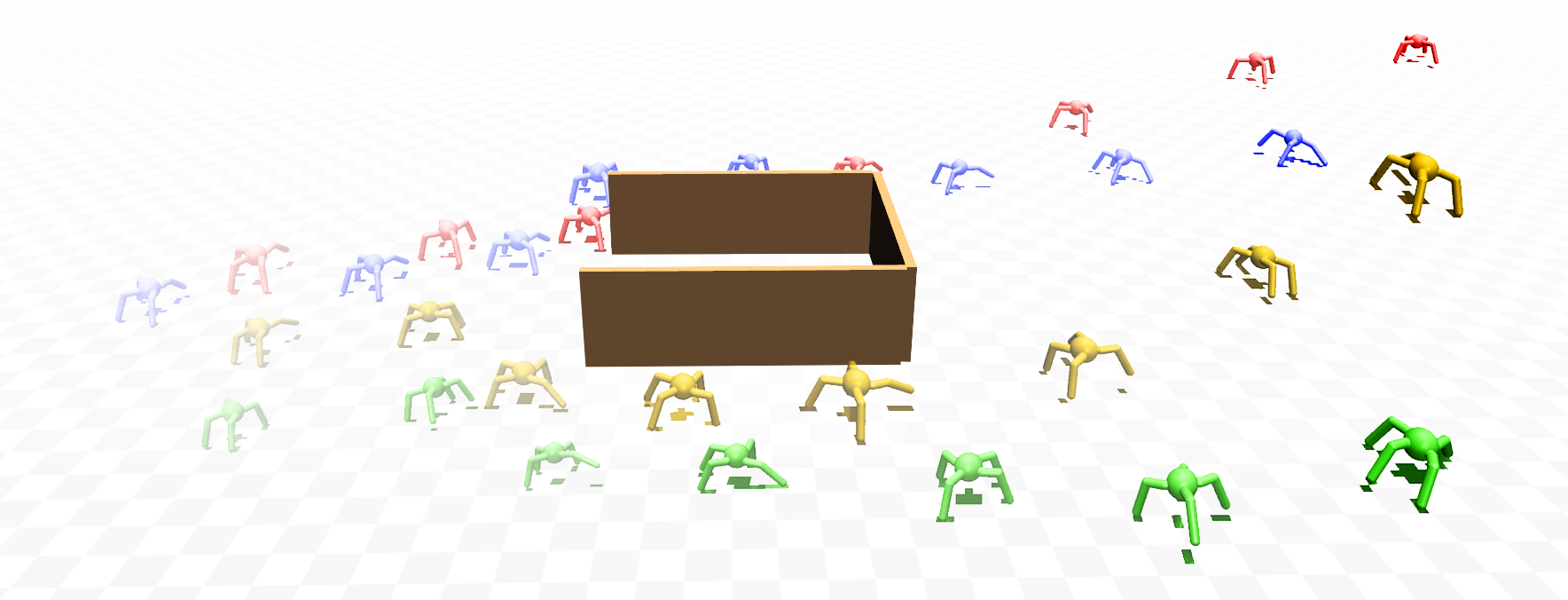}
        \Description{AntMaze}
    \caption{The AntTrap environment: an ant robot learns to run as fast as possible in the forward direction. The trap constitutes a sub-optimal local optimum from which most methods do not manage to escape.}
    \label{fig:teaser}
\end{teaserfigure}

\maketitle

\section{Introduction}

Quality-Diversity (QD) \citep{cully2017quality} is a family of optimization methods based on evolutionary algorithms (EA). Standard optimization methods aim to find an optimal solution that maximizes a criterion, often called fitness in the evolutionnary community. In contrast, QD methods seek a population of diverse solutions that are all high-performing in their niche. A solution is described by a vector of scalar numbers, called solution behavior descriptor, that are used to characterize the novelty of a solution. A solution is considered novel compared to other solutions if its behavior descriptor is different from the descriptors of the other solutions. The difference can be quantified using a distance metric over the behavior descriptor space. \me \citep{mouret2015illuminating}, a popular family of QD methods, splits the behavior descriptor space into a collection of cells and aims to find a solution that maximizes the fitness function in each cell. Such methods have been widely used across multiple domains to efficiently evolve collections of solutions. Having diverse solutions to a single problem is handful in many situations, for example to ease the sim-to-real gap, help for fast adaptation \citep{cully2015robots} or simply to offer the user many possible options.

Classical QD methods rely on divergent random search to optimize both for quality and diversity. While this technique was proved to work in many benchmarks, it is limited by poor scaling capacities preventing its use in applications such as neuro-evolution and control in large search spaces. Recently, a few works \citep{colas2020scaling, Nilsson2021, pierrot2022qdpg} showed that methods like \me can be extended to find good candidate solutions in neuro-evolution applications despite very large search spaces. Notably, \pgame \citep{Nilsson2021} proposed to incorporate policy gradient updates to focus and accelerate the \me search. The authors trained simulated legged robots controlled by neural networks with nearly hundred thousand of parameters to run as fast as possible while obtaining diversity in the robots' gaits.

Most of the mentioned works focus on applications where diversity is a target in itself. Diversity can also be a way to improve exploration in hard control tasks. While the former type of applications has been widely studied, the latter remains scarcer in the QD literature. Exploration is a central theme in many control frameworks such as Reinforcement Learning \citep{sutton2018reinforcement} and remains a hard obstacle for these methods in many situations \citep{amin2021explorl, tang2016exploration, burda2018exploration}. Recent works such as \qdpg \citep{pierrot2022qdpg} demonstrated that QD methods are promising candidates to offer an efficient and general method to solve these exploration difficulties. We believe that hard exploration in control problems involving neural controllers is an exciting research direction for the QD community. In this paper, we propose to highlight and study a set of three benchmarks that fit in that framework and that constitute good candidates for the systematic evaluation of any QD method for neuro-evolution. In addition, we provide open-source code of these environments in \jax \citep{jax2018github}, enabling to run fast and numerous experiments even on limited compute resources.

\section{Proposed benchmarks}
The three highlighted benchmarks are tasks where a controller (sometimes called actor in the reinforcement learning literature) interacts with an environment by taking continuous actions in an environment defined as a Markov Decision Process (MDP). An MDP $\left(\mathcal{S}, \mathcal{A}, \mathcal{R}, \mathcal{T}, \right)$ is described by its state space $\mathcal{S}$, its action space $\mathcal{A}$, its reward function $\mathcal{R}$ and its transition function $\mathcal{T}$. The state space $\mathcal{S}$ corresponds to the set of possible information observed by the controller that can be used to choose actions in the action space $\mathcal{A}$. Once an action is taken in a state, the next state reached in the environment is determined by the dynamics transition function $\mathcal{T}: \mathcal{S} \times \mathcal{A} \rightarrow \mathcal{S}$. Each experienced transition is also associated with a reward determined by function $\mathcal{R} : \mathcal{S} \times \mathcal{A} \rightarrow \mathbb{R}$.

We focus on neuro-evolution, meaning that each controller is defined with a neural network approximator that maps states to continuous actions $\pi_{\theta}: \mathcal{S} \rightarrow \mathcal{A}$. The parameters $\theta$ of the neural network define what we call the genotype of the controller.

In each task, we define the fitness of a controller as the sum of the rewards obtained by the controller along a trajectory. In order to assess the diversity of a controller, we further introduce a behavior descriptor (BD) space $\mathcal{B}$ and a behavior descriptor extraction function $\mathbf{\xi}: \Theta \rightarrow \mathcal{B}$ that characterizes an aspect of the trajectory of the controller in the environment. Finally, every state in the environment can optionally be described by a state descriptor. In this work, we assume that there exist a mapping between state and behavior descriptors such that maximizing diversity in the state descriptor space leads in most cases to diversity in behavior descriptor space. In all the tasks we present, the behavior descriptor of a controller is simply defined as the final state descriptor of the controllers trajectory.

In this paper, we present three tasks: \ptmaze, \antmaze and \anttrap. Such tasks have already been widely used in previous works \citep{parker2020effective, colas2020scaling, frans2017meta, shi2020efficient}. The three tasks focus on evolving controllers in environments defined as MDPs, where the controller's fitness will depend on its capacity to explore the environment.

\begin{figure}[h!] %[thbp!]
 \centering
 \begin{subfigure}{.32\linewidth}
 \centering
   \includegraphics[width=1\linewidth]{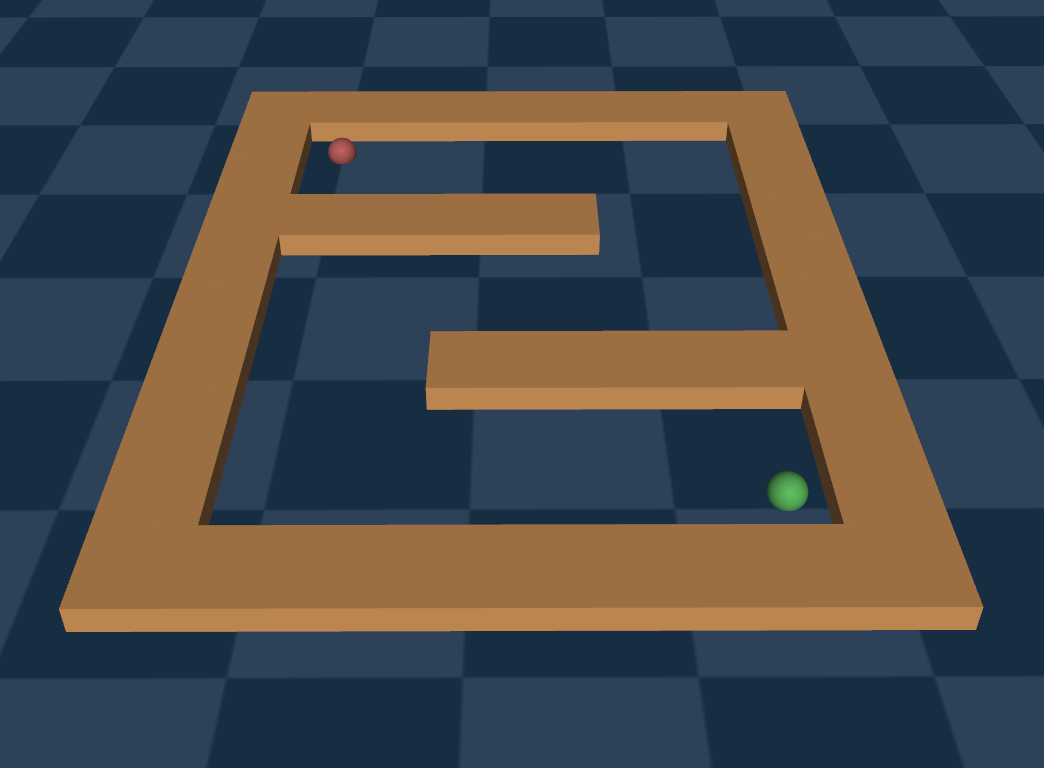}
   \caption{\ptmaze}
   \label{fig:point_maze_inertia}
 \end{subfigure}%
 \hfill
 \begin{subfigure}{.32\linewidth}
 \centering
   \includegraphics[width=1\linewidth]{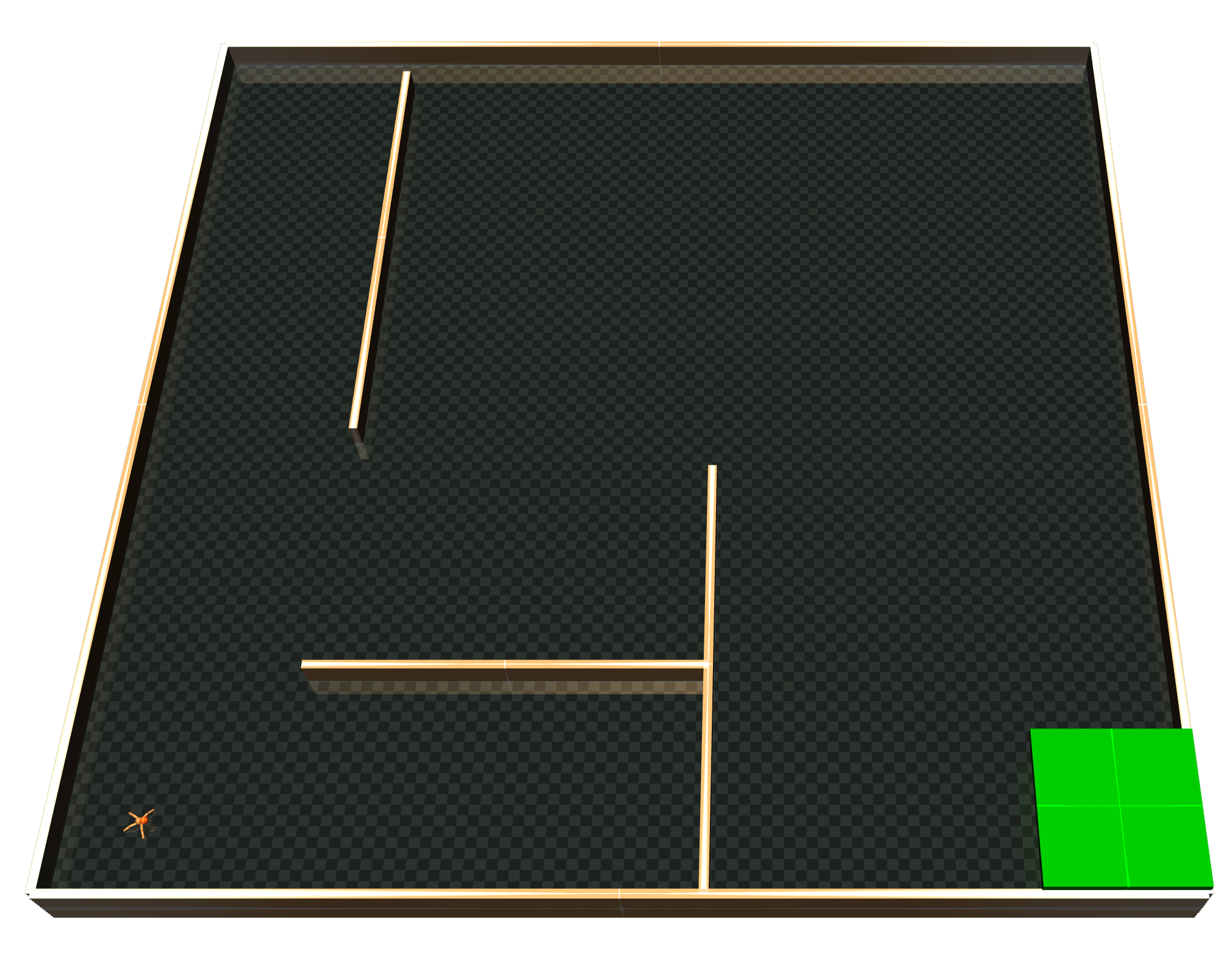}
   \caption{\antmaze}
   \label{fig:ant_maze}
 \end{subfigure}%
 \hfill
 \begin{subfigure}{.32\linewidth}
 \centering
   \includegraphics[width=1\linewidth]{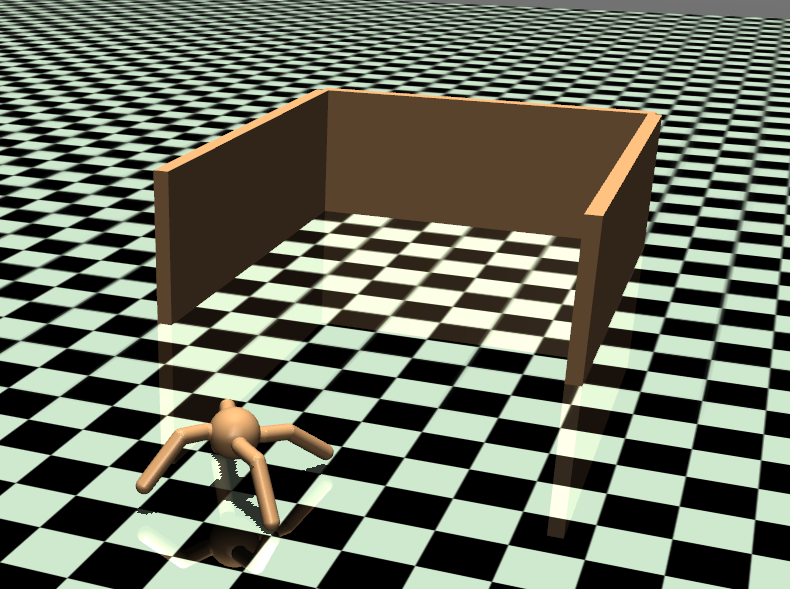}
   \caption{\anttrap}
   \label{fig:ant_trap}
 \end{subfigure}
     \caption{Presented benchmarks \ptmaze, \antmaze and \anttrap.}
     \label{fig:envs}
 \end{figure}

\subsection{\ptmaze}

In \ptmaze, a controller is given 200 time-steps to find the exit of a maze, located in the top left corner, while starting in the bottom right corner. The state observed by the controller is its x/y position, thus a two-dimensional vector, and the action that the controller can take corresponds to an increment of its x/y position, hence a two dimension vector as well. The maze is a $2 \times 2$ square with two walls and the moves amplitude are limited to 0.1 (i.e. the action [1, 1] corresponds to a [0.1, 0.1] move in the maze, unless a wall is hit). The reward at every time-step is defined as the negative Euclidean distance between the controller and the end of the maze. Rewards are typically between -2.5 and -0.1. An episode is stopped when the controller is less than 0.05 away from the exit. 

This environment can be visualised on figure \ref{fig:envs}: the controller is the green sphere and the end of the maze corresponds to the red sphere.

\begin{itemize}
\item {\textbf{genotype space}}: the genotype space equals the possible weights of a neural networks with two hidden layers of size 64. Smaller networks would be expressive enough to learn how to exit the maze but we find interesting to impose a high-dimensional search space.
\item{\textbf{behavior space}}: the behavior descriptor is defined as the x/y position of the point in the maze at the end of an episode. The behavior space is hence $[-1, 1] \times [-1, 1]$.
\item{\textbf{state descriptor space}}: at each step, the state descriptor is defined as the x/y position of the point. The state descriptor space is hence equal to the behavior descriptor space.
\item{\textbf{fitness function}}: the fitness function is defined as the sum of the rewards collected by the agent in the environment.
\end{itemize}

\ptmaze does not include locomotion: the actions directly impact the x/y position in the maze. It is a simple yet efficient way to assess the exploration capacity of an algorithm.

\subsection{\antmaze}

The \antmaze environment is modified from OpenAI Gym \ant \citep{brockman2016openai}, introduced in in \citep{colas2020scaling} and inspired from \citep{frans2017meta}. In \antmaze, a four-legged ant is given 3000 timesteps to reach a goal zone located at the bottom right of the environment, while starting at the bottom left of the maze. At each time-step, the state is the vector containing the angle and angular speed of each joint defining the ant robot body. The action is a 8-dimensional vector describing the torques applied to these joints (two joints for each four legs of the ant).

The reward is defined in the same way as in \ptmaze, it is the negative Euclidean distance between the x/y position of the center of gravity of the ant and the center of the goal area to reach. This maze is a $75 \times 75$ square with walls. It is particularly large and hence hard to solve \citep{vemula2019contrasting}.

This environment can be visualise on figure \ref{fig:ant_maze}, the ant is located at the bottom left and the goal area corresponds to the green area at the bottom right.

\begin{itemize}
\item {\textbf{genotype space}}: the genotype space equals the possible weights of a neural networks with two hidden layers of size 256. Unlike \ptmaze, it is necessary to have this typical size of genotype space to be able to get high fitness in this task.
\item{\textbf{behavior space}}: the behavior descriptor is defined as the x/y position of the center of gravity of the ant at the end of the episode. The behavior space is hence $[-35, 40] \times [-35, 40]$.
\item{\textbf{state descriptor space}}: at each step, the state descriptor is defined as the x/y position of the center of gravity of the ant. The state descriptor space is hence equal to the behavior descriptor space.
\item{\textbf{fitness function}}: the fitness function is defined as the sum of the rewards collected by the agent in the environment.
\end{itemize}

The \antmaze can be seen as an advanced version of the \ptmaze, the exploration problem being enriched with a locomotion challenge: in \ptmaze, the action directly impacts the x/y position, whereas in \antmaze, the action impacts the gait of the Ant, potentially leading to a move in the environment.

\subsection{\anttrap}

The \anttrap environment derives from \ant as well and is inspired from \humanoidtrap, an environment introduced in \citep{conti2018improving} and used in \citep{colas2020scaling, parker2020effective}. 

In \anttrap, a four-legged ant is given 1000 time-steps to run as fast as possible in the x direction in an environment where the ant faces a trap in its initial position.  Just like in \antmaze, the state contains angle and angular speed of the robot joints. The state also contains contact forces of the body parts with the ground and the trap, leading to a 113-dimensional vector. The action is the torque applied to the joints (8 dimensions). At each time-step, the reward is defined as the forward speed of the ant, plus some control costs and survival bonus.

\begin{itemize}
\item {\textbf{genotype space}}: the genotype space equals the possible weights of a neural networks with two hidden layers of size 256.
\item{\textbf{behavior space}}: the behavior descriptor is the x/y position of the center of gravity of the ant at the end of the episode and we clip it in a region of interest. The behavior space is hence $[0, 30] \times [-8, 8]$.
\item{\textbf{state descriptor space}}: at each step, the state descriptor is defined as the x/y position of the center of gravity of the ant, clipped as well. The state descriptor space is equal to the behavior descriptor space.
\item{\textbf{fitness function}}: the fitness function is the sum of the rewards collected in the environment. Running into the trap prevents high speed and stops the ant, which prevents accumulating rewards.
\end{itemize}

The \anttrap task gathers both exploration and locomotion challenges like \antmaze but brings an interesting perspective to the benchmark through the fact that the fitness and behavior descriptor are not aligned. This makes it more difficult for \me and also for pure novelty seeking approaches as simply exploring the behavior descriptor space is not enough to find a performing solution.

\section{Experiments}
\label{sec:exps}

\begin{figure*}[h!]
   \centering
   \includegraphics[width=1\linewidth]{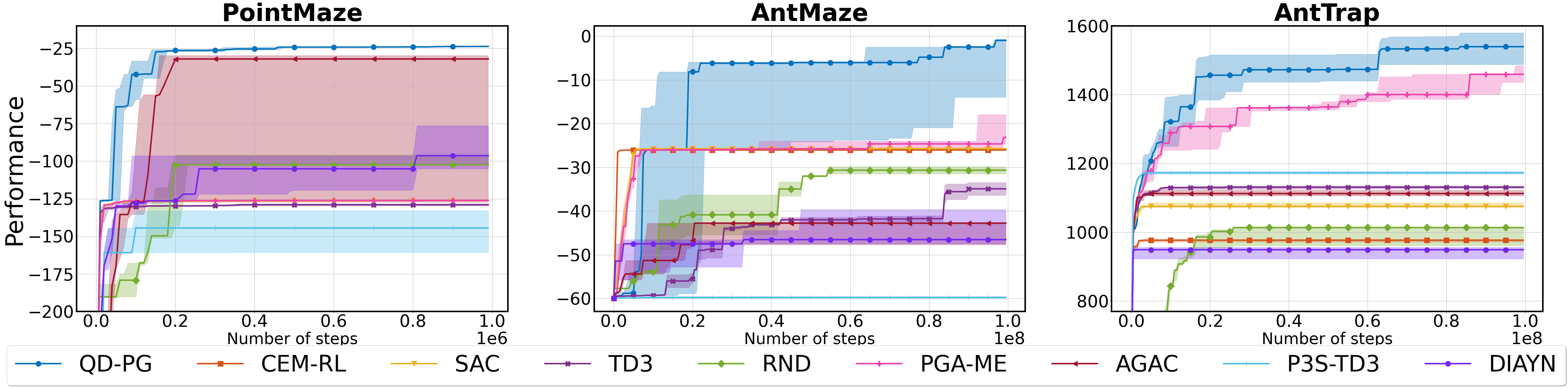}
   \caption{Performance of baseline algorithms for \ptmaze ($10^6$ steps), \antmaze ($10^8$ steps) and \anttrap ($10^8$ steps). Plots present median bounded by first and third quartiles. Figure taken from \citep{pierrot2022qdpg}.}
   \label{fig:performance_results_comparison}
\end{figure*}

\subsection{Baselines and Ablations}

To show the interest of these 3 benchmarks, we present results from \qdpg \citep{pierrot2022qdpg}, comparing many baselines from different fields of continuous control learning algorithms.

These results include QD baselines, namely \mees, \nsres, \nsraes from \citep{colas2020scaling}. These baselines use mutations mechanisms based on Evolutionary Strategies like \citep{salimans2017evolution} to replace divergent random search in a QD loop.
They also include state-of-the-art Reinforcement Learning algorithms, that we describe as pure Policy Gradient baselines. They rely on variants of the Policy Gradient theorem introduced in \citep{silver2014deterministic}. Twin Delayed Deep Deterministic policy gradient (\tddd) \citep{fujimoto2018addressing} is build on the Deterministic Policy Gradient theorem. Soft Actor Critic (\sac) \citep{haarnoja2018soft} is a similar method, but rely on stochastic actions decisions, furthermore, \sac uses an entropy regularization term to help exploration. Random Network Distillation (\rnd) \citep{burda2018exploration} uses the error between a fixed and a learned network to get a measure of the novelty of a state. This error being increased each time a state is visited, hence giving incentives for exploration of novel states. Adversarially Guided Actor Critic (\agac) \citep{AGAC2021} also introduces an exploration incentive to its learning process, through the use of an adversarial network. These methods help assessing the challenge that our benchmark represents for pure policy gradient based search, enriched or not with exploration incentives. 
It is also interesting to assess performances of a pure diversity seeking policy gradient method like \diayn \citep{eysenbach2018diversity}, which seeks diversity by maximizing an objective inspired by information theory. Diversity is learned through a set of skills (similar to behaviors) that visit different states. This approach could hence provide skills (behaviors) that explore the environment.
We consider Population-based Reinforcement Learning methods as well: \cemrl \citep{pourchot2018cem} mixes policy gradient and evolutionary updates towards a reward maximization objective, \pssstddd \citep{jung2020population} trains a set of controllers and constrains diversity in the actions taken by these controllers.
Finally, the baselines contains QD methods using policy gradients: \pgame \citep{Nilsson2021} mixes genetic crossover and policy gradient updates in a \me grid and \qdpg \citep{pierrot2022qdpg} uses two types of policy gradient updates: one seeking for reward accumulation, the other seeking for novel or rare state visitation.\\

Results are gathered in Table~\ref{tab:comp_ablations_rl}. 
Results for \mees, \nsres and \nsraes are only available for \antmaze and were taken from the original paper \citep{colas2020scaling}. All the other results were run with 5 seeds for each baseline.

\subsection{Results}
\label{sec:results}

\begin{table} %[!htb]
    \caption{Final performance of the baselines algorithms. {\bf Final Perf.} is the minimum distance to the goal in \antmaze (after $10^8$ steps) and the best fitness in \ptmaze (after $10^6$ steps) and \anttrap (after $10^8$ steps). Given numbers are medians and mean distances between median and first (resp. third) quartile. Taken from \citep{pierrot2022qdpg}.}
    %\begin{scriptsize}
    %\begin{subtable}[t]{\linewidth}
        \centering
        \label{tab:comp_ablations_rl}
        \begin{tabular}{cccl} %{l|ccc}
            \toprule
            %& \multicolumn{3}{c}{\textbf{Final Perf.} ($\pm$ std)}  \\
            %Final performance ($\pm$ std)\\
            %\midrule
            \textbf{Algorithm} & \ptmaze & \antmaze & \anttrap\\
            \midrule
            \qdpg & ${\bf -24 \pm 0}$ & ${\bf -1 \pm 7}$ & ${\bf 1540 \pm 46}$\\
            \midrule
            \sac & $-126 \pm 0$ & $-26 \pm 0$ & $1075 \pm 7$\\
            \tddd & $-129 \pm 1$ & $-35 \pm 1$ & $1131 \pm 4$\\
            \rnd & $-102 \pm 4$ & $-31 \pm 1$ & $1014 \pm 27$\\
            \midrule
            \cemrl & $-312 \pm 1$ & $-26 \pm 0$ & $977 \pm 3$\\
            \pssstddd & $-144 \pm 14$ & $-60 \pm 0$ & $1173 \pm 4$\\
            \agac & $-32 \pm 49$ & $-43 \pm 3$ & $1113 \pm 8$\\
            \diayn & $-96 \pm 14$ & $-47 \pm 4$ & $949 \pm 34$\\
            \pgame & $-126 \pm 0$ & $-18 \pm 6$ & $1455 \pm 17$\\
            \bottomrule
            %\caption{Comparison to ablations and PG baselines. }
        \end{tabular}
    %\end{subtable}%
\end{table}

\textbf{How challenging are the considered benchmarks for evolutionary methods?} As proved in \citep{Nilsson2021} and \citep{colas2020scaling}, \me with genetic mutations and crossovers is not able to evolve a performing population of controllers on such high-dimensional genotype spaces. On the reported results \ref{tab:comp_evo}, we can see that \me augmented with Evolution Strategies (\mees) is able to eventually reach good performance but with a very low data efficiency. This approach, based on estimations of the natural gradient, is limited by the available resources: it is time-efficient thanks to parallelism and important resources (thousands CPUs). This kind of infrastructure is unfortunately not available for most practitioners.

Given that \me and \mees do not use the structure of the controller (neural network can be differentiated) nor the time-step information available (rewards) that are available in these environments, these methods are limited in terms of final performance or at least data-efficiency that they can achieve.\\

\begin{table}
    %\begin{subtable}[t]{\linewidth}
        \centering
        \caption{Comparison of evolutionary methods on \antmaze. Inspired from \citep{pierrot2022qdpg}.}
        \label{tab:comp_evo}
        \begin{tabular}{ccc} %{l|ccc}
        % & \multicolumn{3}{c}{\antmaze} \\
        \toprule
        \textbf{Algorithm} & \textbf{Final Perf.} & \textbf{Steps to goal}\\
        \midrule
        \qdpg & $-1 \pm 7$ & ${\bf 8.4e7}$\\
        \cemrl & $-26 \pm 0$ & $\infty$\\
        \mees & $-5 \pm 0$ & $2.4e10$\\
        \nsres & $-26 \pm 0$ & $\infty$\\
        \nsraes & ${\bf -2 \pm 1}$ & $2.1 e10$\\
        \bottomrule
        \end{tabular}
    %\end{subtable}
    %\end{scriptsize}
\end{table}

\textbf{How challenging are the considered benchmarks for pure policy gradients methods?} Table~\ref{tab:comp_ablations_rl} presents performances of state-of-the-art policy gradient algorithms and show that pure policy gradient methods cannot find high-performing solutions in these benchmarks. \tddd quickly converges to local minima in all the environments. Entropy regularization in \sac does not help to do better than \tddd, this mechanism does not provide enough exploration to escape the local minima induced by the misleading rewards.
 
The exploration intrinsic reward used in \rnd was reported to help fully explore \ptmaze in some seeds but was too brittle to get high fitness on all the five seeds and manage to handle the most challenging exploration environment \antmaze and \anttrap.

\diayn shows exploration capacity in \ptmaze but not enough to explore the whole maze. An issue with \diayn is that once the skills are different enough to be discriminated, there is no more incentive to explore: hence, if the learned skills can be discriminated while reaching only states from the first part of the maze, they will be no incentive to try to reach a deeper part of the maze. Moreover, as soon as the behavior descriptor is not aligned with the fitness, like in \anttrap, \diayn produces very low fitness controllers, as there is no pressure for performance.

\agac is the second best performing baseline on \ptmaze but suffers from a very high variance, and could not scale well to the other environments, probably suffering from the increase of the action space dimension. The last presented method trying to maintain diversity in the action space, \pssstddd, got low results in the 3 environments.

Finally, the exploration mechanism based on \cem used in \cemrl is also unable to provide enough exploration. The controllers trained by this algorithm converges to local optima in the three presented benchmarks.\\

\textbf{How are the best methods performing on these tasks?}

We can see a clear gap between \qdpg, \pgame and the other baselines in this benchmark. These two methods are the only able to get more than 1200 in \anttrap and to pass -25 in \antmaze (with a clear advantage for \qdpg on \antmaze, able to score a median of -1). Even if the final best fitness obtained by \pgame is still far from the goal area, scoring above -25 shows that the best final controller had been able to go around all the walls. Nevertheless, we can see that \pgame cannot solve \ptmaze within the $10^6$ time-steps limit, showing that the genetic crossovers are lacking of data-efficiency when exploration is challenging and when the genotype space is high-dimensional. Finally, \qdpg is able to get strong performance on all three benchmarks but figure \ref{fig:performance_results_comparison} shows that it suffers from a higher variance on \antmaze. Furthermore, \citep{pierrot2022qdpg} acknowledges that \qdpg struggles on gait locomotion tasks from \citep{Nilsson2021}, because it is difficult to link the descriptors at time-step and trajectory level in these tasks. Table \ref{tab:env_difficulty} summarize the properties of the presented baselines that seemed crucial to solve the tasks.

This study of ten algorithms from the literature shows that these three tasks are definitely challenging and we believe that the best performances obtained by \qdpg could still be improved in terms of data-efficiency. Furthermore, as the time of writing, getting high fitness in the three presented tasks in addition to the locomotion benchmarks presented in \citep{Nilsson2021, flageat2022pgameanalysis} remains an open challenge.

\begin{table}
        \centering
        \caption{We compare properties of some of the studied algorithms and their ability to solve or not the three proposed benchmarks. QPG (resp. DPG) stands for Quality (resp. Diversity) Policy Gradient. PG stands for Reinforcement Learning methods that do not rely on evolution.}
        \label{tab:env_difficulty}
        \begin{tabular}{cccc} %{l|ccc}
        \toprule
        \textbf{Environment} & \textbf{PG} & \textbf{QD + QPG} & \textbf{QD + QPG + DPG}\\
        \midrule
        \ptmaze & X & X & \checkmark\\
        \antmaze & X & X & \checkmark\\
        \anttrap & X & \checkmark & \checkmark\\
        \bottomrule
        \end{tabular}

\end{table}

\section{Jax implementation of the benchmark}

\begin{table}
        \centering
        \caption{Performance comparison of \mujoco and \brax implementations of the presented environments. Represented values are steps per second on a single CPU. We give the mean and standard deviation for each value.}
        \label{tab:time_perf}
        \begin{tabular}{cccc} %{l|ccc}
        % & \multicolumn{3}{c}{\antmaze} \\
        \toprule
        \textbf{Implem.} & \textbf{\ptmaze} & \textbf{\antmaze} & \textbf{\anttrap}\\
        \midrule
        \mujoco/NumPy & $9820 \pm 180$ & ${1170 \pm 20}$ & ${1470 \pm 50}$\\
        \brax/\jax & $(1.52 \pm 0.13)e6$ & ${4480 \pm 110}$ & ${7470 \pm 180}$\\
        \bottomrule
        \end{tabular}
\end{table}

Google has recently introduced \jax \citep{jax2018github}, a package based on XLA that enables high-performance numerical computation in Python. \jax enables to compile code and to run it on hardware accelerators. As pointed out by \citep{lim2022qdax}, this a great tool for Quality Diversity methods, as they can take full advantage of the vectorization of many operations. Amongst others, the evaluations of the controllers in the environments can now be compiled and executed on hardware accelerators like GPU (or even TPU) thanks to a fully \jax-implemented physical simulator \brax \citep{freeman2021brax}. \citep{lim2022qdax} shows impressive improvements of \me run-time compared to previous implementations, even outperforming pure C++ implementations for batch sizes larger than 500 (and being 30 times faster for batch size of $10^5$) on a GPU.

We hence decided to use \brax to re-implement the environments presented in this work and made them publicly available in QDax~\citep{chalumeau2023qdax}. \ptmaze, that has no physical simulations, is implemented in \jax and follows \brax interface. To confirm the interest of this new implementation, we launch episodes of equal time duration of the \mujoco and \brax versions of the environments.\\

Table \ref{tab:time_perf} reports steps per second of our implementations when running on one CPU. With our Python/NumPy version of \ptmaze, we can launch 10000 steps per second, versus more than 1.5 millions steps per second with the \jax compiled version, resulting in a 150 speed-up factor. On one CPU, the \jax/\brax version of \anttrap (resp. \antmaze) is performing 5 (resp. 4) times more steps than the \mujoco version.

Furthermore, whereas it is hard to scale with \mujoco implementations, requiring use of tools like MPI to manage multiple processes, and impossible to make use of the hardware accelerator, the use of \texttt{\jax.vmap} is all it takes to batch the environment rollouts on the GPU. Although there is an initial loss of performance when running a single environment on GPU (compared to CPU), we experience a quasi-constant run-time to launch 1 versus 1000 parallel rollouts in our \brax implementations, using a single Tesla T4 GPU. Results reported on table \ref{tab:gpu_perf}.

\begin{table}
    %\begin{subtable}[t]{\linewidth}
        \centering
        \caption{Time steps per second on a GPU with several batch sizes, running our \jax/\brax implementations.}
        \label{tab:gpu_perf}
        \begin{tabular}{cccc} %{l|ccc}
        \toprule
        \textbf{Batch size} & \textbf{\ptmaze} & \textbf{\antmaze} & \textbf{\anttrap}\\
        \midrule
        1 & $21260 \pm 1860$ & ${345 \pm 3}$ & ${356 \pm 3}$\\
        10 & $(2.14 \pm 0.22)e5$ & ${3290 \pm 20}$ & ${3330 \pm 30}$\\
        100 & $(2.14 \pm 0.14)e6$ & ${31520 \pm 90}$ & ${30840 \pm 80}$\\
        1000 & $(2.03 \pm 0.27)e7$ & ${1.77e5 \pm 740}$ & ${(2.3 \pm 0.014)e5}$\\
        \bottomrule
        \end{tabular}
    %\end{subtable}
    %\end{scriptsize}
\end{table}

\section{Conclusion}

We presented three control tasks that exhibit both a high dimensional search space and exploration difficulties. The results reported in this paper show that pure evolutionary methods and pure policy gradient methods struggle in these benchmarks and confirms the interest of the approaches mixing both. In addition, we show how these environments are complementary and hence convenient for a broader understanding of the studied algorithms, with different level of alignment between behavior descriptors and fitness. A \jax implementation of these benchmarks is made open-source. We think that these hard exploration benchmarks, put aside the locomotion benchmarks introduced in \citep{Nilsson2021}, build a coherent and complete additional set of benchmarks that can be used to assess exploration performances of Quality Diversity algorithms for neuro-evolution.


\begin{acks}

Work by Nicolas Perrin-Gilbert was partially supported by the French National Research Agency (ANR), Project ANR-18-CE33-0005 HUSKI.

\end{acks}

\bibliographystyle{ACM-Reference-Format}
\bibliography{qdexplo}

\end{document}